\documentclass[10pt,twocolumn,letterpaper]{article}

\usepackage{cvpr}
\usepackage{times}
\usepackage{epsfig}
\usepackage{graphicx}
\usepackage{amsmath}
\usepackage{amssymb}
\usepackage{diagbox} 
\usepackage[]{footmisc}
\usepackage{multirow}

\usepackage[pagebackref=true,breaklinks=true,letterpaper=true,colorlinks,bookmarks=false]{hyperref}

\cvprfinalcopy 

\def\cvprPaperID{7346} 

\ifcvprfinal\fi
\begin{document}

\title{Hierarchical Feature Embedding for Attribute Recognition}

\author{Jie Yang\textsuperscript{1,2}, 
Jiarou Fan\textsuperscript{1}, 
Yiru Wang\textsuperscript{1},
Yige Wang\textsuperscript{1}, 
Weihao Gan \textsuperscript{1}\thanks{Corresponding author} , 
Lin Liu\textsuperscript{2}, 
Wei Wu\textsuperscript{1}\\
\textsuperscript{1}SenseTime Group Limited, \textsuperscript{2}Tsinghua University\\
{\tt\small takanashiyj@gmail.com,\{fanjiarou,wangyiru,ganweihao,wuwei\}@sensetime.com}, \\
{\tt\small yige.wang@tum.de,linliu@tsinghua.edu.cn}
\and
}

\maketitle
\thispagestyle{empty}

\begin{abstract}
   Attribute recognition is a crucial but challenging task due to viewpoint changes, illumination variations and appearance diversities, etc. Most of previous work only consider the attribute-level feature embedding, which might perform poorly in complicated heterogeneous conditions. To address this problem, we propose a hierarchical feature embedding (HFE) framework, which learns a fine-grained feature embedding by combining attribute and ID information. In HFE, we maintain the inter-class and intra-class feature embedding simultaneously. Not only samples with the same attribute but also samples with the same ID are gathered more closely, which could restrict the feature embedding of visually hard samples with regard to attributes and improve the robustness to variant conditions. We establish this hierarchical structure by utilizing HFE loss consisted of attribute-level and ID-level constraints. We also introduce an absolute boundary regularization and a dynamic loss weight as supplementary components to help build up the feature embedding. Experiments show that our method achieves the state-of-the-art results on two pedestrian attribute datasets and a facial attribute dataset.
\end{abstract}

\section{Introduction}

Attributes, such as gender, hair length, clothing style, are discriminative semantic descriptors that can be used as soft-biometrics in visual surveillance. Attribute recognition concentrates on discerning these attributes of the target human in a given image. It includes pedestrian attribute recognition (PAR), face attribute recognition (FAR), etc. Recently, attribute recognition has received extraordinary attention owing to the potential applications in person re-identification (Re-ID) \cite{layne2012person,liu2012person,peng2016joint}, face verification \cite{kumar2009attribute, chopra2005learning, taigman2014deepface, nguyen2010cosine, sun2013hybrid}, and human identification \cite{jaha2014soft}. Being a classification problem by nature, it still faces great challenges in real-world scenarios for these reasons: (1) Images might be low-resolution or blurry because of the shot distance or the movements of pedestrians. (2) Different scenes, time slots, angles and poses lead to illumination alterations, viewpoint changes, and appearance variations. (3) Some parts of an object might be occluded by others, resulting in invisibility or ambiguity. 

\begin{figure}[t]
\centering
\includegraphics[width=0.45\textwidth]{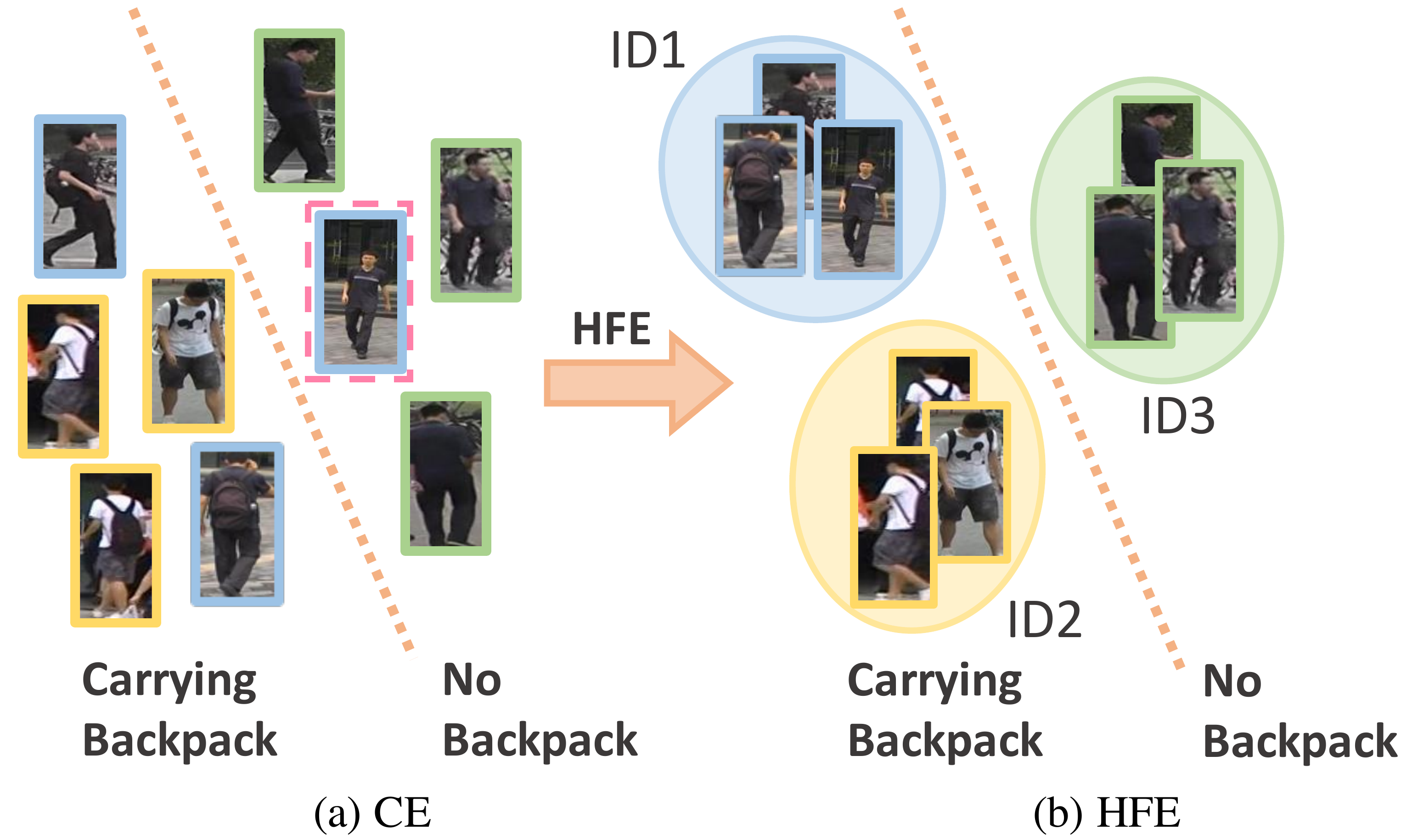}
\caption{Hierarchical feature embedding on `backpack' attribute. Images with the same IDs are represented with the same color borders. (a) represents the Cross Entropy feature space, where most of samples can be classified correctly while the hard sample with pink dotted border (whose backpack is totally occluded by the body) is misclassified. (b) With the help of ID constraint, features with the same IDs form fine-grained clusters can pull the hard example back.}
\label{intro}
\end{figure}

Recently, some methods are introduced to solve these problems and achieve admirable performance. A-AOG \cite{park2017attribute} explicitly represents the decomposition and articulation of body parts, and accounts for the correlations between poses and attributes. LGNet \cite{liu2018localization} assigns attribute-specific weights to local features based on the affinity between pre-extracted proposals and attribute locations. These methods aim at applying pivotal parts of images or an attention module to reduce the impact of irrelevant factors to some extent, still they do not cope with the visual appearance variations of attributes as well as occlusion directly. Only attribute-level optimization is focused on in these methods, however the information from attribute recognition related fields, such as person Re-ID, could assist to alleviate variation and occlusion issues through imposing stronger constraints.

From the perspective of data, current attribute datasets are all labelled on ID or tracking granularity \cite{lin2019improving,liu2015deep} to reduce workloads. Thus we assume that images captured from the same identity should have the same attributes, but not vise versa. For each attribute, labels are usually coarse-grained owing to the expensive annotation cost. Different persons may get the same attribute labels but with subtle differences in appearance. For example, backpacks with different colors and textures are all tagged as `backpack'. Therefore, fine-grained feature embeddings are needed for attributes to represent the variety within a class. With identity information, we could set up a two-level feature embedding, i.e., inter-class and intra-class (Fig. \ref{intro}). For each attribute, samples with identical attribute form coarse-grained classes, while in each attribute class, samples from the same person (with the same attribute definitely) construct the fine-grained ID clusters. We introduce this hierarchical feature embedding model by the following motivations: (1) ID clusters restrict the images with the same ID but variant viewpoints, illumination and appearance to gather more closely, which embed the attribute features with scene invariance and improve robustness. (2) Hard cases for attributes may be easily handled and pulled back by other easy samples of the same ID by the ID constraint, which is difficult to learn only in the attribute level. (3) Like the attribute tags, the ID tags are also utilized in the attribute semantics by holding the assumption above, avoiding integrating different semantic features directly in the same feature space as previous work \cite{lin2019improving}.

Motivated by the above observation, we propose a hierarchical feature embedding (HFE) framework, maintaining the inter-class as well as the intra-class feature embedding by combining attribute and ID information. HFE loss is introduced for fine-grained feature embedding, which consists of two triplet losses and an absolute boundary regularization with the selected quintuplets. With HFE loss constraints, each class could gather more compactly, leading to a more distinct boundary between classes. We propose the absolute boundary regularization for additional absolute constraints because the triplet loss only considers the difference between two distances but ignores the absolute values, and the intra-class triplet loss may interact indirectly with the inter-class boundary. Besides, the quintuplet selection is relevant to the current feature space. However, in the early stage of training, feature spaces are not confident enough for the quintuplet selection, so we design a dynamic loss weight, making the HFE loss weight increasing gradually along with the learning process. In summary, the contributions of this paper are:

\begin{itemize}
\item We propose HFE framework to integrate ID information on attribute semantics for fine-grained feature embedding. A novel HFE loss is introduced for both inter-class and intra-class level constraints.
\item We construct an absolute boundary regularization by strengthening the original triplet loss with an absolute constraint.
\item We introduce a dynamic loss weight, which forces the feature space to transit from origin to the improved HFE-restricted space by degrees.
\item The proposed method is evaluated on two pedestrian attribute datasets and one face attribute dataset. Experiments show our method achieves the state-of-the-art results on all three datasets. 
\end{itemize}

\section{Related Work}

\begin{figure*}[t]
\centering
\includegraphics[width=0.85\textwidth]{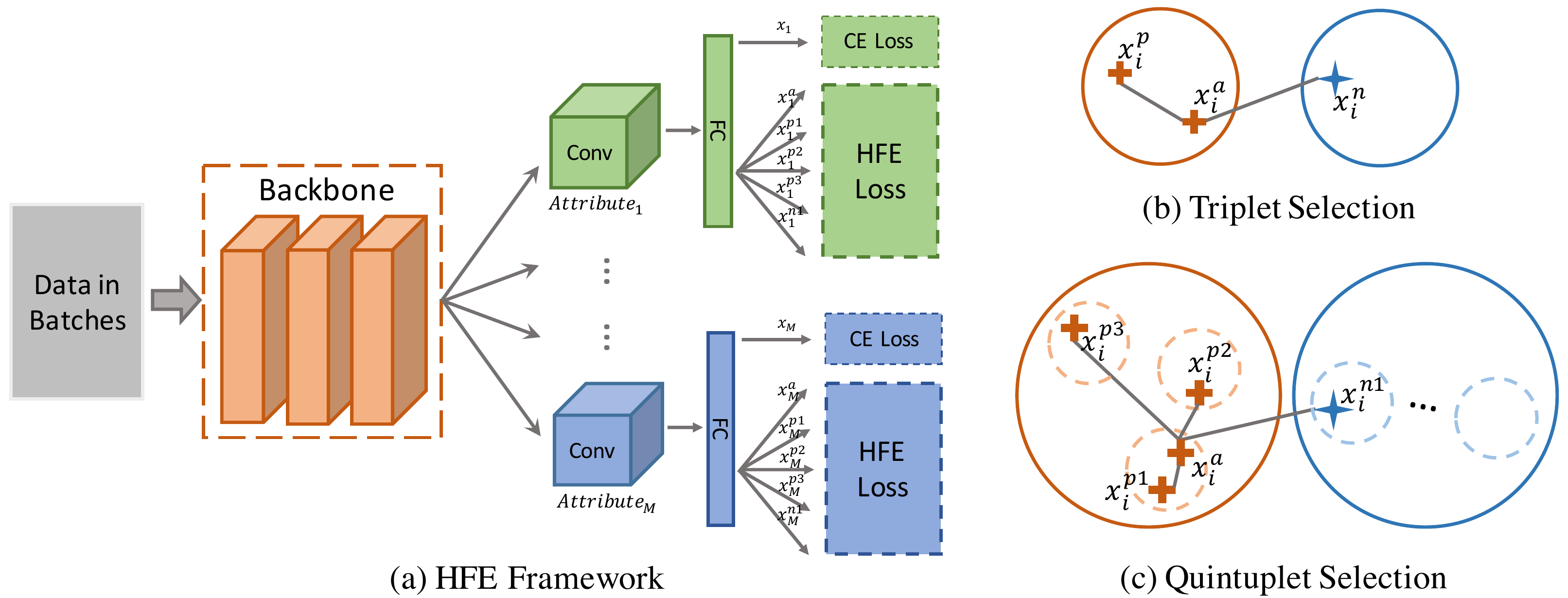}
\caption{(a) Overview of the proposed hierarchical feature embedding (HFE) framework, it consists of a backbone model, attached by $M$ branches for $M$ attributes. We calculate the CE loss and HFE loss based on the quintuplet selection in each branch. (b) and (c) are triplet and quintuplet selection respectively, where orange and blue represent the different attribute classes.}
\label{fig1}
\end{figure*}
\subsection{Attribute Recognition}
 Recently, deep learning based attribute recognition methods achieve impressive performance. In PAR, these methods includes global based \cite{li2015deepmar, sudowe2015person, abdulnabi2015multi, deng2014pedestrian}, local parts based \cite{liu2018localization}, visual attention based \cite{liu2017hydraplus}, sequential prediction based \cite{zhao2018grouping} methods, etc. Among them, DeepMar \cite{li2015deepmar} is an early global based PAR work. Considering the imbalanced data distribution, it proposes cost-sensitive cross entropy (CE) loss for classification. It also proposes a new loss to handle imbalanced data as well as a new attention mechanism. LGNet \cite{liu2018localization} assigns attribute-specific weights to local features based on the affinity between pre-extracted proposals and attribute locations. Hydraplus-Net \cite{liu2017hydraplus} proposes an attention based model and exploits the global and local contents with multi-level feature fusion of a single pedestrian image. ALM \cite{tang2019improving} aims at learning attribute-aware representation through attribute localization. Attribute-aware attention model \cite{han2018attribute} exploits the correlation between global features and attribute-specific features and utilize it to generate attention mask in a reciprocal manner. Localizing by describing \cite{liu2017localizing} learns the bounding boxes related to the location of attributes explicitly using REINFORCE algorithm with a designed reward function in a weakly supervised manner. \cite{sarafianos2018deep} designs an attention mechanism for aggregating multi-scale features as well as a loss function similar to focal loss \cite{lin2017focal} in order to tackle the imbalanced data problem. GRL \cite{zhao2018grouping} proposes a RNN based grouping recurrent learning method that exploits the intra-group mutual exclusion and inter-group correlation. FAR can be also divided into part-based and holistic approaches \cite{rudd2016moon,he2018harnessing}.

ReID aims at matching a target person in a set of query pedestrian images. A great amount of deep learning based ReID works provide promising solutions \cite{guangcong2019aaai,fu2019horizontal,chen2019abd, ahmed2015improved,Gong_ICCV,Qin_2020_CVPR}. A bunch number of existing methods rely on exploiting discriminative features, which is in the same spirit as fine-grained recognition. Attributes and ReID are highly correlated pedestrian visual appearance representations, but vary in semantics and granularities. Even though they are different tasks, the common characteristic can be beneficial to both of them, which is exploiting discriminative features. And therefore they are reasonable to be dealt with jointly. Some works utilize both information for multi-task learning \cite{sun2018unified} or assisting the main task \cite{lin2019improving,liu2018sequence}. The methods can be summarized in two categories: (1) shared backbone and task-independent branches (2) task-independent models and combining high level features in some way (e.g., concatenated FC). For example, APR \cite{lin2019improving} learns ReID embedding and pedestrian attributes simultaneously, sharing the same backbone and owning classification FC layers respectively. UF \cite{sun2018unified} trains two different models for two tasks and concatenates branches to one identity vector for ReID. These methods combine these two kind of features to some extent. However incorporating them into the coefficient feature representation indiscriminately is less powerful since attribute recognition and ReID are essentially divergent. Persons with similar attributes can also be different identities. Therefore a more rational way of combining both information is required.

\subsection{Metric Learning}

The objective of metric learning \cite{yi2014deep, khan2016unsupervised} is to learn a proper metric feature space so that the distances between similar samples reduce and that of dissimilar samples enlarge. While traditional metric learning algorithms \cite{kulis2013metric} are based on linear transformation, nonlinear models, etc, due to the recent advances in deep learning, Convolutional Neural Networks have been a powerful tool to learn a task-specific metric and achieved impressive results in a wide range of tasks. Many metric learning algorithms have been proposed in image retrieval \cite{wang2014learning}, ReID \cite{chen2017beyond}, face recognition \cite{schroff2015facenet, sun2015deepid3, liu2017sphereface,Wu_2020_CVPR}, etc. Representative methods are contrastive loss and triplet loss. Contrastive loss \cite{hadsell2006dimensionality} restricts pair inputs and results in distances between similar pairs as close as possible and that of dissimilar pairs to be larger than margin. Triplet loss \cite{schroff2015facenet} applies triplet as input and ensures the difference between the distance of (anchor, negative) feature and (anchor, positive) feature is larger than margin. Beyond triplet loss, quadruplet loss \cite{chen2017beyond} and quintuplet loss \cite{huang2016learning} are also introduced to improve performance. Center loss \cite{wen2016discriminative} is designed for face recognition strives to push samples to their respective clusters centers. 
We propose HFE loss by applying inter-class and intra-class triplet losses for fine-grained constraints.

\section{Proposed Method}
\textbf{Problem Definition.} Given $N$ images $\{I_1, I_2,...,I_N\}$ and each image $I_j$ has $M$ visual attribute tags $y_j=\{y_{j1}, y_{j2},...,y_{jM}\}$ together with ReID label $l_j$. The images from the same person are labelled with identical attributes, i.e., $\forall_{l_i=l_j} y_i=y_j$. ReID auxiliary attribute recognition aims at training a model containing the attribute and ID information to predict the attributes $y_k$ for the characteristic of the person in an unseen image $I_k$.

\textbf{Network Architecture.} As shown in Fig. \ref{fig1} (a), the proposed hierarchical feature embedding (HFE) network consists of a backbone model, to which $M$ branches for $M$ attributes are attached. In the shared backbone, the model learns a common feature embedding for all attributes. For each attribute, we construct branches respectively for two reasons: (1) Different attributes, such as age and gender, should own their specific feature embeddings. (2) We construct metric loss for each attribute in their own feature spaces, which can not be applied on a shared feature space. For example, there are two images $I_1,I_2$ from different IDs, and the attributes are (long hair, carrying backpack) and (long hair, no backpack). We should pull them closer for hair length feature while push them away for backpack feature. Each attribute branch contains Conv-BN-ReLU-Pooling-FC sequential layers. We calculate the Cross Entropy (CE) loss and metric loss on each branch.

\textbf{Loss Computation.} We apply CE loss for attribute classification (Eq. \ref{ce_loss}) as most works do. Besides, an HFE loss is utilized for auxiliary metric learning with weight $w$ (Eq. \ref{all_loss}). HFE loss consists of inter-triplet loss, intra-triplet loss and Absolute Boundary Regularization, which will be introduced in the next section.

\begin{equation}
L_{CE}=-\frac{1}{N}\sum_{i=1}^{N}\sum_{j=1}^{M} y_{ij}log(p_{ij})+(1-y_{ij})log(1-p_{ij})
\label{ce_loss}
\end{equation}

\begin{equation}
Loss=L_{CE}+wL_{HFE}
\label{all_loss}
\end{equation}

\subsection{Hierarchical Feature Embedding}\label{section:hfe}
\textbf{Triplet Loss}. Triplet loss has been widely used for metric learning. As shown in Fig. \ref{fig1} (b), it trains on a series of triplets $\{x_i^a,x_i^p,x_i^n\}$, where $x_i^a$ and $x_i^p$ are image features from the same label, and $x_i^n$ from a different label. $a$, $p$ and $n$ are abbreviations of anchor, positive and negative sample respectively. The formulation is as follows:

\begin{equation}
L_{trp}=\frac{1}{N}\sum_{i=1}^{N}[d(x_i^a,x_i^p)-d(x_i^a,x_i^n)+\alpha]_+
\label{eq1}
\end{equation}

Here, $d(.)$ represents the Euclidean distance, and $\alpha$ is the margin which forces the gap of $d(x_i^a,x_i^n)$ and $d(x_i^a,x_i^p)$ larger than $\alpha$. $[z]_+$ means $max(z,0)$. When the gap is larger than $\alpha$, the triplet loss would be zero. 

\textbf{Inter-class Triplet Loss.} We can extend triplet loss to attribute classification scenario. As shown in Eq. \ref{inter}, $x_{ij}^a$ is the feature of anchor sample $I_i$ of $j$-th attribute, associated with the feature of a positive and negative sample in regard to $x_{ij}^a$, i.e, $x_{ij}^{p3}, x_{ij}^{n1}$. Here we define the triplet on the attribute level, and $\alpha_1$ is the inter-class margin.

\begin{equation}
\begin{aligned}
L_{inter}=&\frac{1}{N}\sum_{i=1}^{N}\sum_{j=1}^{M}[d(x_{ij}^a,x_{ij}^{p_3})-d(x_{ij}^a,x_{ij}^{n_1})+\alpha_1]_+ \\
&y_{ij}^{p_3}=y_{ij}^{a},\  y_{ij}^{n_1}\not=y_{ij}^{a},\ l_{ij}^{p_3}\not=l_{ij}^{a},\  l_{ij}^{n_1}\not=l_{ij}^{a}
\end{aligned}
\label{inter}
\end{equation}
We use batch hard mode \cite{schroff2015facenet} for triplet selection. In each batch, we take the closest negative sample to anchor as the hard negative sample $x_{ij}^{n_1}=argmin_{x_{ij}^n}d(x_{ij}^a,x_{ij}^n)$, and the farthest positive as the hard positive sample $x_{ij}^{p_3}=argmax_{x_{ij}^p}d(x_{ij}^a,x_{ij}^p)$, which is called as the hard inter-class triplet loss.

\textbf{Intra-class Triplet Loss.} With the attribute inter-class triplet loss, we could separate the feature embedding between classes. However, the feature embeddings in each class are still mixed up. Intuitively, the features of samples with similar appearance or the same ID should be closer than others. However, it is not easy to get such a perfect feature embedding without extra constraint on the intra-class level. To form ordered and fine-grained intra-class feature embeddings, we utilize ID information to enforce the features that belong to the same person gathered more closely. We construct the intra-class feature embedding for these two reasons: (1) The intra-class triplet loss restricts the features from the same person to gather more closely, making the embedding more robust to scene variance. (2) The hard cases for attributes but not for ID can be easily handled by ID clusters in the attribute feature embedding. Here we introduce a hard intra-class triplet loss, similar to the hard inter-class triplet loss, while the hard negative sample is replaced by the closest positive sample to the anchor with different ID but the same attribute, $x_{ij}^{p_2}=argmin_{x_{ij}^p}d(x_{ij}^a,x_{ij}^p)$ for $y_{ij}^{p}=y_{ij}^a,\ l_{ij}^{p}\not=l_{ij}^a$, and the hard positive sample is turned into the farthest positive sample with the same ID (with the same attribute definitely), $x_{ij}^{p_1}=argmax_{x_{ij}^p}d(x_{ij}^a,x_{ij}^p)$ for $y_{ij}^{p}=y_{ij}^a,\ l_{ij}^{p}=l_{ij}^a$. The intra-class triplet loss is shown in Eq. \ref{eq2}, and $\alpha_2$ is the intra-class margin. 

\begin{equation}
\begin{aligned}
L_{intra}=&\frac{1}{N}\sum_{i=1}^{N}\sum_{j=1}^{M}[d(x_{ij}^a,x_{ij}^{p_1})-d(x_{ij}^a,x_{ij}^{p_2})+\alpha_2]_+ \\
&y_{ij}^{p_1}=y_{ij}^a,\  y_{ij}^{p_2}=y_{ij}^a,\ 
l_{ij}^{p_1}=l_{ij}^a,\  l_{ij}^{p_2}\not=l_{ij}^a
\label{eq2}
\end{aligned}
\end{equation} 

To maintain the structures of inter-class and intra-class feature embedding simultaneously, we incorporate the inter-class and intra-class triplet loss.
As shown in Fig. \ref{fig1} (c), HFE loss takes quintuplet samples $\{x_{ij}^a,x_{ij}^{p_1},x_{ij}^{p_2},x_{ij}^{p_3},x_{ij}^{n_1}\}$ as input and manage to maintain the multiple  relative relationships, $d(x_{ij}^a,x_{ij}^{p_1}) < d(x_{ij}^a,x_{ij}^{p_2}) <d(x_{ij}^a,x_{ij}^{p_3}) <d(x_{ij}^a,x_{ij}^{n}) $. The quintuplet is summarized in Table \ref{tb:quintuplet}. With the constraints on both inter-class and intra-class level, we can construct a hierarchical feature embedding to incorporate attribute information as well as ID information in the attribute feature space.

\begin{table}[]
	\centering
	\resizebox{0.35\textwidth}{!}{
		\begin{tabular}{l|c|c|c}
		\hline
			item & attribute & ID & distance \\ \hline
			$x_{ij}^{p_1}$ & the same & the same & farthest \\
			$x_{ij}^{p_2}$ & the same & different & nearest \\
			$x_{ij}^{p_3}$ & the same & different  & farthest \\
			$x_{ij}^{n}$ & different  & different  & nearest \\ \hline
		\end{tabular}

	}
	\vspace{3pt}
	\caption{The summary of quintuplet. $x_{ij}^a$ is the anchor and not listed above.}
	\label{tb:quintuplet}
\end{table}

\subsection{Absolute Boundary Regularization}

Triplet loss only restricts the difference between $d(x_{ij}^a,x_{ij}^{p})$ and $d(x_{ij}^a,x_{ij}^{n})$ while ignores the absolute value. The difference is dependent on the selected triplet in batch, which is hard to ensure $d(x_{ij}^a,x_{ij}^{p}) < d(x_{ij}^a,x_{ij}^{n})$ in the whole training dataset.

In our constraints, to guarantee attributes owing discriminative intra-class feature embeddings, we pull away $x_{ij}^{p_2}$ from $x_{ij}^a$ relative to $x_{ij}^{p_1}$. Although margin $\alpha_2 < \alpha_1$, $L_{intra}$ may interact indirectly with inter-class boundary. 

Considering these two factors, we force $d(x_{ij}^a,x_{ij}^{n})$ to be larger than an absolute distance $\alpha_3$, named as Absolute Boundary Regularization (ABR).

\begin{equation}
L_{ABR}=\frac{1}{N}\sum_{i=1}^{N}\sum_{j=1}^{M}[\alpha_3-d(x_{ij}^a,x_{ij}^n)]_+
\end{equation}

We compose these three parts in Eq. \ref{eq4} as HFE loss, which aims at a hierarchical feature embedding, considering not only inter-class features but also inner-class, not only relative distance but also absolute to ensure the boundary more discriminative.
\begin{equation}
L_{HFE}=L_{inter}+L_{intra}+L_{ABR}
\label{eq4}
\end{equation}

\subsection{Dynamic Loss Weight}

 In the early stage of training, the feature space is not good enough for quintuplet selection. Therefore, applying a large weight for HFE loss at the beginning may amplify the noise caused by the initial model. To solve this problem, we set a small weight for HFE loss at the beginning, making the original CE loss play a major role in optimization and produce an elementary feature space. Then we enlarge the weight afterwards to refine the original feature space to be more fine-grained. So we introduce a dynamic loss weight, modifying the composite function proposed by \cite{wang2019dynamic} to control the weight increasing from small to large nonlinearly. In Eq. \ref{eq5}, $T$ means the total training iterations, and $iter$ means the current iteration. $w_0$ is a given value.
 
\begin{equation}
w=[\frac{1}{2}cos(\frac{T-iter}{T}\pi)+\frac{1}{2}]w_0
\label{eq5}
\end{equation}
In the training process, we increase the HFE loss weight from zero to the given value, forcing the feature space to transit from origin to the improved HFE-restricted space by degrees. 

\section{Experiment}

\begin{table*}[t]
	\centering
	\resizebox{1\textwidth}{!}
	{
		\begin{tabular}{l|cccccccccccc|c}
			\hline
			Method                      & gender    & hair  &  L.slv &      L.low &    S.clth & hat & B.pack& bag & H.bag & age & C.up & C.low& avg.      \\ \hline
			CE &  84.86 & 84.51 &  93.49 & 90.64 &  90.24 & 96.01 & 79.58 & 74.16 &  86.67 & 92.62 & 92.99 & 92.86 & 88.22\\
			DeepMAR & 85.24&85.48&92.79&91.37&89.37&95.93&84.56&71.57&86.53&\textbf{94.56}&93.78&92.92&88.68 \\
			\hline
		    
			APR & 85.64 &	85.62	&92.87&	92.80&	89.91&	97.13&	78.12	&75.41	&\underline{90.53}&	93.18&	93.18	&92.77&	88.93 \\
			UF &88.94&	78.26&	93.53	&92.11&	84.79	&97.06&	85.46	&67.28&	88.4&	84.76&	87.5&	87.21&	86.28\\ 
			JCM &89.7&82.5&\underline{93.7}&\textbf{93.3}&89.2&\underline{97.2}&85.2&\textbf{86.9}&86.2&87.4&92.4&93.1&89.73 \\ \hline
			HP Net & \underline{94.74}&89.11&93.14&92.47&92.06&96.94&\underline{87.33}&79.65&89.22&93.28&93.21&92.39&91.13 \\
			DIAC    & 93.32&\underline{90.43}&93.24&92.82&\textbf{95.63}&96.98&86.18&77.56&88.84&92.88&\textbf{94.55}&\underline{92.84}&\underline{91.27} \\ \hline
			HFE (Ours) & \textbf{94.88}&\textbf{90.51}&\textbf{94.03}&\underline{93.25}&\underline{94.18}&\textbf{97.88}&\textbf{90.41}&\underline{85.35}&\textbf{91.45}&\underline{94.37}&\underline{94.43}&\textbf{94.00}&\textbf{92.90}\\ \hline
			
		\end{tabular}

	}
	\vspace{3pt}
	\caption{Class-based evaluation on Market 1501 attribute dataset with the best results in bold and the second best results underlined. `L.slv', `L.low', `S.clth', `B.pack', `H.bag', `C.up', `C.low' denote length of sleeve, length of lower-body clothing, style of clothing, backpack, handbag, color of upper-body clothing and color of lower-body clothing, respectively. }
	\label{tb:market_acc}
\end{table*}

\begin{table*}[t]
	\centering
	\resizebox{0.9\textwidth}{!}
	{
		\begin{tabular}{l|cccccccccc|c}
			\hline
			Method                      & gender    & L.up  &  boots &      hat &    B.pack &bag & H.bag & C.shoes & C.up & C.low & avg.      \\ \hline
			CE &  82.33&86.63&88.36&82.98&73.31&80.65&91.60&90.92&95.40&91.45&86.36 \\
			DeepMAR &82.26&87.14&88.49&82.15&75.84&82.54&91.53&\underline{91.38}&95.18&92.52&86.90 \\
			\hline
		    
			APR &83.47&87.44&88.02&86.98&75.79&82.16&92.61&90.67&94.23&\underline{97.43}&87.88 \\
			UF &\textbf{88.94}&\textbf{93.6}&80.13&82.97&87.02&\underline{91.60}&89.60&83.65&93.94&91.84&88.33 \\ 
			JCM &\underline{87.4}&88.3&89.6&83.3&\textbf{89.0}&87.9&92.4&87.1&92.9&92.1&89.00\\ \hline
			HP Net &83.91&87.58&86.72&78.91&77.54&83.37&93.40&88.91&\textbf{96.81}&97.19&87.43 \\
			DIAC &85.87&89.74&\underline{89.63}&\textbf{90.79}&82.90&87.88&\underline{93.47}&90.21&\underline{95.92}&97.11&\underline{90.35} \\ \hline
		
			HFE (Ours) &87.02&\underline{89.88}&\textbf{90.70}&\underline{88.69}&\underline{88.50}&\textbf{91.81}&\textbf{93.64}&\textbf{93.82}&95.85&\textbf{97.80}&\textbf{91.77} \\ \hline
			
		\end{tabular}

	}
	\vspace{3pt}
	\caption{Class-based evaluation on Duke attribute dataset with the best results in bold and the second best results underlined. `L.up', `B.pack', `H.bag', `C.shoes', `C.up', `C.low' denote length of sleeve, backpack, handbag, color of shoes, color of upper-body clothing and color of lower-body clothing, respectively. }
	\label{tb:duke_acc}
\end{table*}

\begin{table*}[]
	\centering
	\resizebox{0.65\textwidth}{!}
	{
		\begin{tabular}{l|cccc||cccc}
			\hline
			
			\multirow{2}{*}{\diagbox{Method}{Metric}} & 
			\multicolumn{4}{c||}{Market 1501 attribute dataset} & \multicolumn{4}{c}{Duke attribute dataset}\\ \cline{2-9} 
			                         & acc.  &  prec. &      recall &    F1& acc.  &  prec. &      recall &    F1       \\ \hline
			
			CE     & 69.21	& 82.55	& 78.43	& 80.44&69.00&81.39&77.24&79.26 \\
			DeepMAR    &69.65&82.60&80.24&81.40&70.67&	81.82&	82.24	&82.03\\
			APR & 70.25&83.52&78.96&81.18&70.10&82.74&79.02&80.83 \\
			HP Net    &74.82&85.26&\underline{83.31}&84.27&67.63&82.77&75.19&79.79\\
			DIAC     &\underline{75.03}&\underline{85.64}&83.18&\underline{84.39}&\underline{74.02}&\underline{84.85}&\underline{83.44}&\underline{83.14} \\ \hline
			HFE    & \textbf{78.01}& \textbf{87.41}& \textbf{85.65}& \textbf{86.52}&\textbf{76.68}&\textbf{86.37}&\textbf{84.40}&\textbf{85.37} \\
		          \hline
		\end{tabular}

	}
	\vspace{3pt}
	\caption{Instance-based evaluation on Market 1501 attribute dataset and Duke attribute dataset with the best results in bold and the second best results underlined.}
	\label{tb:instance_based}
\end{table*}

\begin{table}[]
	\centering
	\resizebox{0.5\textwidth}{!}{
		\begin{tabular}{l|c|c|c|c|c}
			\hline
			Metric Loss                      & avg.    & acc.  &  prec. &      recall &    F1       \\ \hline
			$None$  & 88.82 & 70.03	& 83.15 & 	79.27	& 81.12	\\
		    $L_{inter}$    &90.83&71.83 	&85.35	&80.63&	82.93	\\
			$L_{intra}$    &91.27&74.61&	85.43&	83.56&	84.48 \\
			$L_{inter}+L_{intra}$     & 92.44					& 77.08                    & 86.73         & 84.88           & 85.99 \\
			$L_{inter}+L_{intra}+ABR$    & 92.73					& 77.57                    & 87.00          & 85.45           & 86.22 \\
			$L_{inter}+L_{intra}+ABR$*     & \textbf{92.90}					& \textbf{78.01}                    & \textbf{87.41 }         & \textbf{85.65}           & \textbf{86.52} \\\hline
			$L_{inter}+L_{pairwise\_intra}$ & 92.30& 76.49& 86.21& 84.53&85.36 \\\hline
		\end{tabular}

	}
	\vspace{3pt}
	\caption{Ablation study on Market 1501 attribute dataset. * means replacing the fixed loss weight with dynamic setting. }
	\label{tb:ablation}
\end{table}

\subsection{Datasets}
\textbf{Market 1501 attribute dataset} \cite{lin2019improving} is an extension of Market-1501 dataset \cite{zheng2015scalable} with person attribute annotations. It contains 32,668 annotated bounding boxes of 1,501 identities and 12 different types of attribute annotations for each identity. Attributes include 10 binary attribute (such as gender, hair length and sleeve length) and 3 multi-class attributes (i.e., age, upper clothing color and lower clothing color). Images are captured from six cameras and each annotated identity is present in at least two cameras.

\textbf{Duke attribute dataset} \cite{lin2019improving} is an extension of DukeMTMC-ReID dataset \cite{zheng2017unlabeled} with person attribute annotations. It contains 36,411 bounding boxes of 1,404 identities and 10 different types of attribute annotations for each identity. Attributes includes 8 binary attribute (such as gender, length of upper-body clothing and wearing boots) and 2 multi-class color attributes for upper clothing and lower clothing.  Images are captured from eight cameras and each annotated identity is present in at least two cameras.

\textbf{CelebA} \cite{liu2015deep} is a large-scale face attribute dataset with annotations of 40 binary classifications (such as eyeglasses, bangs and pointly nose). The dataset contains 202,599 images from 10,177 identities and covers large pose variations and background clutter.

\subsection{Evaluation}
For the first two PAR datasets, we evaluate attribute recognition performance on both class-based and instance-based level. (1) \textbf{Class-based:} We calculate the classification accuracy for each attribute class and report the mean accuracy of all attributes \cite{lin2019improving}. (2) \textbf{Instance-based:} We measure the accuracy, precision, recall and F1 score for all test samples. For accuracy, precision and recall, we first compute the scores of predicted attributes against the gound truth for each test sample image and then get the average scores over all test cases. The F1 score is computed based on precision and recall \cite{li2015deepmar}. The gallery images are used as test set and we transform the multi-class attributes into binary classes \cite{lin2019improving}. 

For face attribute dataset, we evaluate class-based mean accuracy. \cite{he2018harnessing}
\subsection{Implementation Details}
The common settings for PAR and FAR: We use Adam \cite{kingma2014adam} as an optimization algorithm. The weight decay is 5e-4, batch size is 256. We random sample 64 identities with 4 images for each to form a batch. Horizontal flip is applied during the training process. We set $\alpha1$, $\alpha2$, $\alpha3$ and $w_0$  with 0.3, 0.1, 5, and 1 respectively.

For PAR, we exploit ResNet 50 as the backbone. The base learning rate is 2e-4 and decays exponentially after epoch 50. We train 130 epochs in total.

For FAR, we exploit DeepID2 \cite{sun2014deep} as the backbone. To accommodate with backbone, in each attribute branch, we replace the convolutional layers with linear layers. The base learning rate is 1e-2 and we use cosine annealing schedule \cite{loshchilov2016sgdr}. We train 300 epochs in total.

\subsection{Experiments on Pedestrian Attribute Dataset}
 We list the results of state-of-the-art methods on these two pedestrian attribute datasets, i.e., Market 1501 attribute dataset and Duke attribute dataset. Table \ref{tb:market_acc} and Table \ref{tb:duke_acc} show class-based metrics and Table \ref{tb:instance_based} indicate instance-based evaluations. Among the compared methods, CE means applying CE loss for attribute classification with the same backbone as ours. DeepMAR\cite{li2015deepmar} applies weighted CE loss function. APR\cite{lin2019improving}, UF \cite{sun2018unified} and JCM\cite{liu2018sequence} are three methods combining attibute and ID information. HP Net\cite{liu2017hydraplus} and DIAC \cite{sarafianos2018deep} are attribute-focused methods that achieve competitive performance recently.
 
 With 8 Titan XP GPUs, our method costs 73.6 seconds per epoch on average for Duke dataset while the CE loss costs 61.2 seconds. Extra time consumption is only on training phase, and no additional computation is needed for inferring.
 
 With regard to attribute and ID joint methods, performances are improved to some extent, but are still inferior to CE for some attributes, such as `backpack', `boots' for APR as well as `hair', `color  of  shoes' for UF and JCM. It shows that combining attribute recognition and ReID in a shared feature space directly is harmful for some attributes. These attributes may not contribute to ReID, resulting in fewer attention by combining ReID directly. However, our method utilizes the ID tags in the semantics of attributes to build up a better feature embedding for attributes, therefore surpassing previous joint methods significantly. Compared with attribute-focused methods, our method achieves better results on highly identity-related attributes, such as `gender' and `age', as well as variant and subtle attributes, such as `bag' and `boots', with the extra ID information.

 Overall, Our method achieves the best performance on both datasets in five evaluation metrics, outperforming the second best results by 1.63\%, 2.98\%, 1.77\%, 2.34\%, 2.13\% on Market 1501 attribute dataset, and 1.42\%, 2.66\%, 1.52\%, 0.96\%, 2.23\% on Duke attribute dataset in average accuracy, accuracy, precision, recall and F1 respectively.
 
\begin{figure*}[t]
\centering
\includegraphics[width=0.75\textwidth]{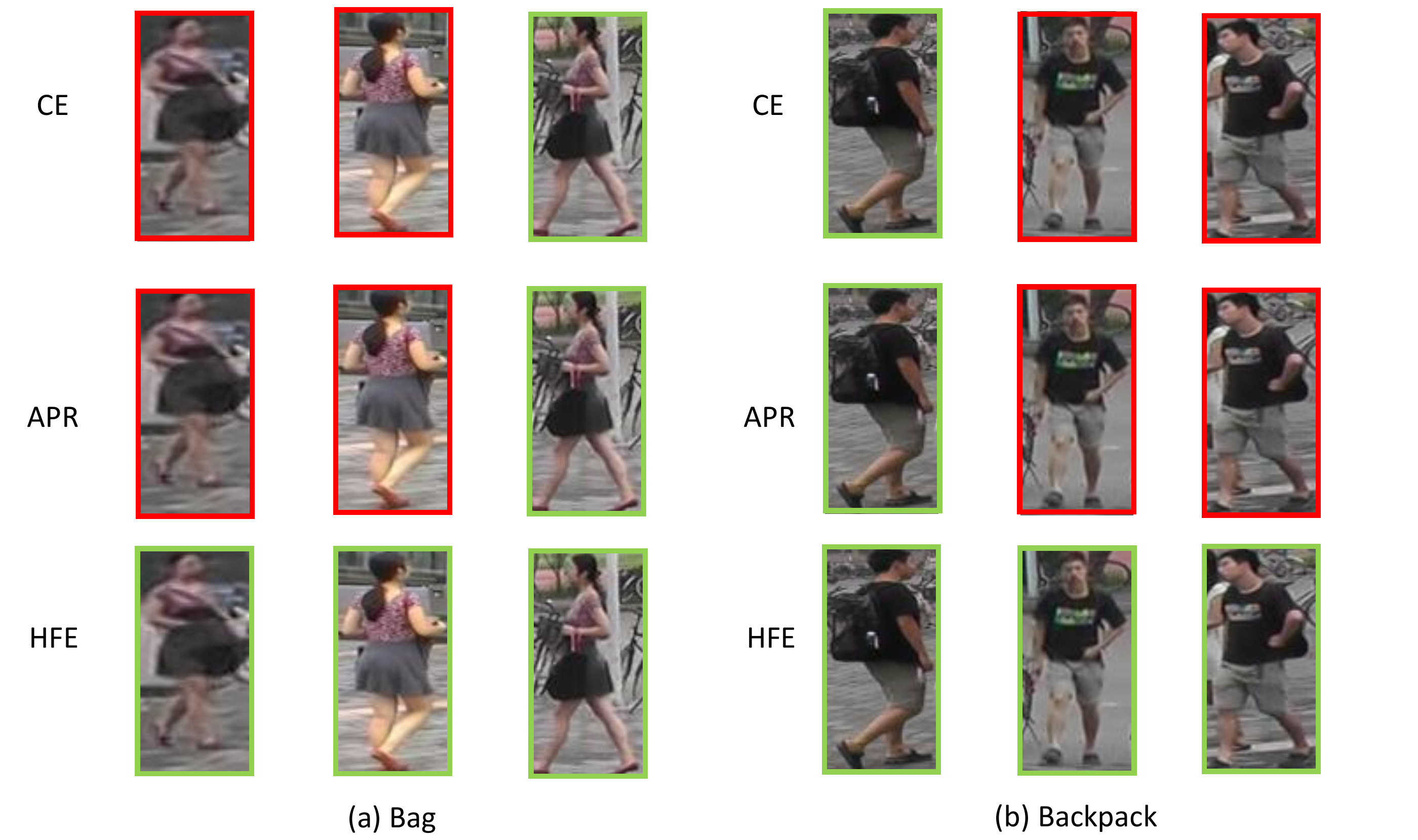}
\caption{Visualization of recognition results. Correct predictions are bounded by green box and red box otherwise. (a) and (b) are prediction results of two IDs on bag and backpack respectively, each compared by three methods.}
\label{fig2}
\end{figure*}

For a more intuitive analysis, we demonstrate the recognition results for two IDs with three images each in Fig. \ref{fig2}. HFE achieves better performance for most attributes, especially for accessories such as bag and backpack, which are sensitive to angle and pose variations. When the attribute object is clearly visible, all three methods achieve good performance. However, when occlusion happens, HFE still predicts correctly thanks to the ID constraint, while the other two methods perform rather poorly. 


\begin{figure}[t]
\centering
\includegraphics[width=0.45\textwidth]{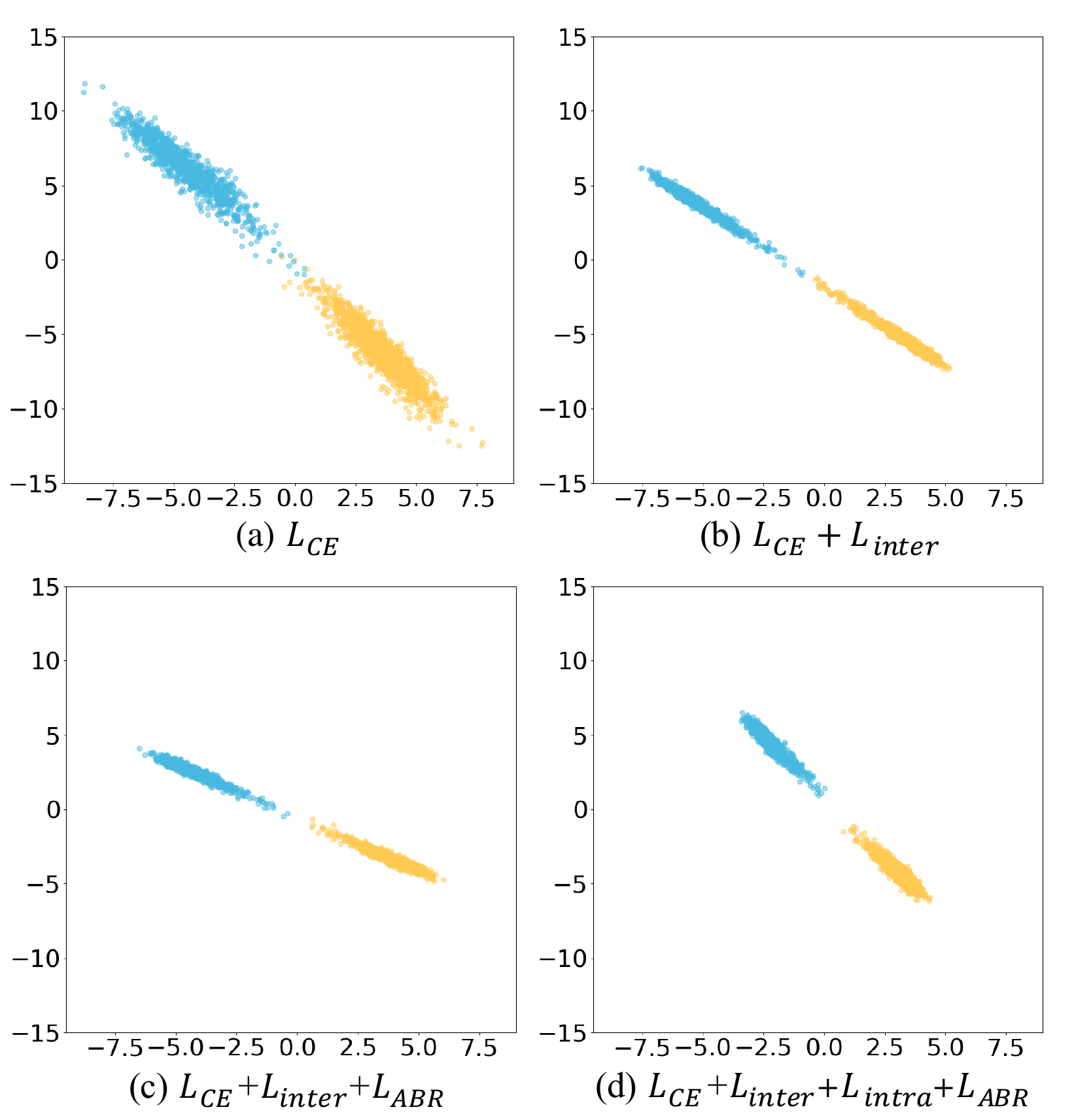}
\caption{Feature embedding visualization of attribute `gender'. Blue points represent `female' and yellow points represent `male'. Figures are in the same scale for fair comparison.}
\label{visual}
\end{figure}
\begin{figure}[t]
\centering
\includegraphics[width=0.46\textwidth]{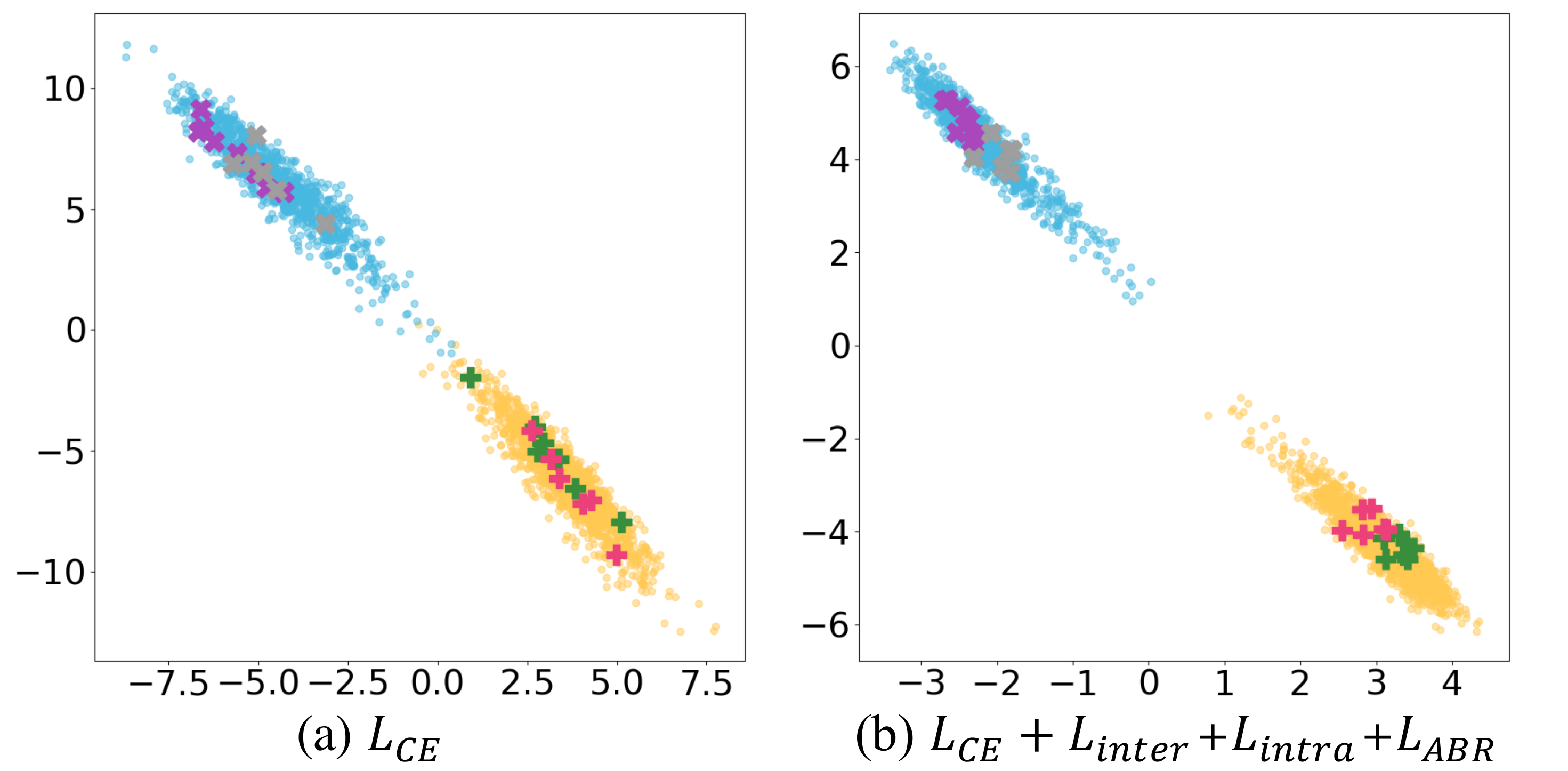}
\caption{ID clusters are visualized for two methods. Four IDs are shown. `+' and `x' markers with the same colors come from the identical ID.}
\label{id_visual}
\end{figure}

\subsection{Ablation Study}
 The advantage of HFE is its capability of learning a fine-grained comprehensive attribute feature representation involving identity information. To better illustrate this, we analyze the effectiveness of each component with quantitative comparisons and qualitative visualization. The ablation study is done on market 1501 attribute dataset.

\textbf{Quantitative Evaluation.} Table \ref{tb:ablation} quantifies the benefits of our hierarchical feature embedding, absolute boundary regularization (ABR), and dynamic loss weight respectively. Based on the CE method, the performance improves distinctly with only $L_{inter}$. While for $L_{intra}$, the improvement is more significant than $L_{inter}$, indicating the extra ID clusters assist attribute classification indeed. Combining $L_{inter}$ and $L_{intra}$ achieves better performance than each of them separately,  demonstrating the complementarity of $L_{inter}$ and $L_{intra}$. The combination achieves 3.62\%, 7.05\% and 4.87\% compared with the CE loss method for avg., acc. and F1 respectively. The improvements of ABR and the dynamic loss weight are also demonstrated in table \ref{tb:ablation}. Finally, with combining all the components, HFE achieves 4.08\%, 7.98\% and 5.40\% improvements for avg., acc. and F1 respectively in total.

In order to verify the necessity of setting up intra-class discriminative embedding by the triplet loss, we replace $L_{inter}$ with a simple pairwise loss which only considers IDs' intra-class compactness but no separability with different IDs. As a result, the triplet loss achieves sightly better performance, which means the inter-class separability of ID is helpful for a fined-grained feature embedding and maintaining detailed attribute information. 

\textbf{Qualitative Evaluation.} We proceed with qualitative evaluation in both inter-class and intra-class level. Fig. \ref{visual} demonstrates the learnt feature embedding visualization of attribute `gender' for different loss function. The CE loss produces two attribute clusters but the boundary is still in mess. Obviously, applying $L_{inter}$ makes the clusters more compact and the margin between classes more distinct. Besides, adding the constraint of $L_{ABR}$, the effect is more prominent. Furthermore, with adding $L_{intra}$, the two classes are restricted by both inter-class and intra-class constraints, leading to tighter clusters and the most discriminating boundary.

Besides, we evaluate the ID clustering ability and effect. As Fig. \ref{id_visual} shows, CE does not control the intra-class arrangements so the IDs are disorganized, whereas HFE can form fine-grained intra-class clusters by the constraint of $L_{intra}$. We can see that from introducing the intra-class constraint, each class gathers more closely and the margin between them is much more distinct, so hierarchical feature embedding does help to classify attributes.
\subsection{Experiments on Face Attribute Dataset}

To evaluate the generalization ability of our framework, we conduct the experiment on a face attribute dataset CelebA. Other state-of-the-art methods are FaceTracer \cite{kumar2008facetracer}, PANDA-l \cite{bourdev2016pose}, LNets+ANet \cite{liu2015deep}, Walk-and-Learn \cite{wang2016walk}, Rudd et al. Moon \cite{rudd2016moon}, CLMLE \cite{huang2019deep}, SSP + SSG \cite{kalayeh2017improving} and HSAI \cite{he2018harnessing} respectively. Table \ref{tb:celeba} shows HFE achieves 92.1\% accuracy and outperforms all other methods. With the help of face ID information, HFE could also build up a fine-grained feature embedding for face attributes, indicating our HFE framework can be easily generalized to similar scenarios, providing a general framework for fine-grained recognition.

\begin{table}[]
	\centering
	\resizebox{0.23\textwidth}{!}
	{
		\begin{tabular}{l|c}
			\hline
			Method       & acc.       \\ \hline
			FaceTracer     & 81.12	     \\ 
			PANDA-l & 85.00 \\
			LNets+ANet & 87.30 \\
			Walk-and-Learn& 88.00\\
			Rudd et al. Moon & 90.94 \\
			CLMLE & 91.13 \\
			SSP + SSG & 91.80 \\
			HSAI & 91.81\\ \hline
			HFE& 92.17 \\
			\hline
		\end{tabular}

	}
	\vspace{3pt}
	\caption{Class-based accuracy on CelebA. }
	\label{tb:celeba}
	\end{table}
	
\section{Conclusion}
In this paper, we present a novel end-to-end Hierarchical Feature Embedding (HFE) framework to explore the combination of attribute and ID information in attribute semantics for attribute recognition. In HFE, each attribute-class is discriminative by the inter-class constraint. Moreover, with the supplementary ID information, we maintain the ID clusters in each attribute class by the intra-class constraint for a fine-grained feature embedding. We apply ID information in attribute semantics and refrain from combining attribute and ID information in the same feature space directly. Furthermore, our mechanism introduce a coarse-to-fine process for discriminative fine-grained feature learning. In addition, we introduce an Absolute Boundary Regularization for combining relative and absolute distance constraint. We also design a dynamic loss weight to force the feature space transiting from the origin to the improved HFE-restricted space by degrees, facilitating the performance of our model and the stability of training. Extensive ablation studies and experimental evaluations justify  effectiveness of our proposed method.

\section{Acknowlegements}
Partial financial support by Tsinghua-Baidu collaborative project and Mercari are gratefully acknowledged.
{\small
\bibliographystyle{ieee_fullname}
\bibliography{egbib}
}

\end{document}